\pdfoutput=1

\documentclass[11pt]{article}

\usepackage[]{EACL2023}

\usepackage{times}
\usepackage{latexsym}

\usepackage[T1]{fontenc}

\usepackage[utf8]{inputenc}

\usepackage{microtype}

\usepackage{inconsolata}

\usepackage{lipsum}  
\usepackage{todonotes}
\usepackage{xcolor}

\usepackage{multirow}
\usepackage{amsmath}
\usepackage{booktabs}
\usepackage{graphicx}
\usepackage{array}
\usepackage{xspace}
\usepackage{hyperref}

\newcommand{\texto}[1]{{\fontfamily{bch}\selectfont{{#1}}}}
\newcommand{\neuroprompts}{\texto{NeuroPrompts}\xspace}

\definecolor{mycolor}{RGB}{0,128,0}
\definecolor{mycolor3}{HTML}{EC008C}

\title{\neuroprompts: An Adaptive Framework to Optimize Prompts for Text-to-Image Generation}

\newcommand{\aspace}{\hspace{1em}}

\author{
    Shachar Rosenman\aspace
    Vasudev Lal\aspace 
    Phillip Howard\aspace \\
    Intel Labs \\
    \texttt{\{shachar.rosenman,vasudev.lal,phillip.r.howard\}@intel.com} \\}

\begin{document}
\maketitle
\begin{abstract}

Despite impressive recent advances in text-to-image diffusion models, obtaining high-quality images often requires \textit{prompt engineering} by humans who have developed expertise in using them. In this work, we present \neuroprompts, an adaptive framework that automatically enhances a user's prompt to improve the quality of generations produced by text-to-image models. Our framework utilizes constrained text decoding with a pre-trained language model that has been adapted to generate prompts similar to those produced by human prompt engineers. This approach enables higher-quality text-to-image generations and provides user control over stylistic features via constraint set specification. We demonstrate the utility of our framework by creating an interactive application for prompt enhancement and image generation using Stable Diffusion. Additionally, we conduct experiments utilizing a large dataset of human-engineered prompts for text-to-image generation and show that our approach automatically produces enhanced prompts that result in superior image quality. We make our 
\href{https://github.com/IntelLabs/multimodal_cognitive_ai/tree/main/Demos/NeuroPrompts}{code}\footnote{\url{https://github.com/IntelLabs/multimodal_cognitive_ai/tree/main/Demos/NeuroPrompts}} and a \href{https://youtu.be/Cmca_RWYn2g}{screencast video demo}\footnote{\url{https://youtu.be/Cmca_RWYn2g}} of \neuroprompts publicly available.



\end{abstract}

\section{Introduction}

Text-to-image generation has recently become increasingly popular as advances in latent diffusion models have enabled widespread use.
However, these models are sensitive to perturbations of the prompt used to describe the desired image, motivating the development of \textit{prompt engineering} expertise by users to increase the quality of the resulting images generated by the model. 

Prompt design is crucial in ensuring that the model accurately comprehends the user's intent.  Text-to-image models face a significant challenge in this aspect as their text encoders have limited capacity,  which can make it difficult to produce aesthetically pleasing images.
Additionally, as empirical studies have shown, common user input may not be enough to produce satisfactory results. Therefore, developing innovative techniques to optimize prompt design for these models is crucial to improving their generation quality.

To address this challenge, we introduce \neuroprompts, a novel framework which automatically optimizes user-provided prompts for text-to-image generation models. A key advantage of our framework is its ability to automatically adapt a user's natural description of an image to the prompting style which optimizes the quality of generations produced by diffusion models. We achieve this automatic adaptation through the use of a language model trained with Proximal Policy Optimization (PPO) \citep{schulman2017proximal} to generate text in the style commonly used by human prompt engineers. This results in higher quality images which are more aesthetically pleasing, as the prompts are automatically optimized for the diffusion model. Furthermore, our approach allows the user to maintain creative control over the prompt enhancement process via constrained generation with Neurologic Decoding \cite{lu-etal-2021-neurologic}, which enables more personalized and diverse image generations.

Our \neuroprompts framework is integrated with Stable Diffusion \citep{rombach2022high} in an interactive application for text-to-image generation. Given a user-provided prompt, our application automatically optimizes it similar to expert human prompt engineers, while also providing an interface to control attributes such as style, format, and artistic similarity.
The optimized prompt produced by our framework is then used to generate an image with Stable Diffusion, which is presented to the user along with the optimized prompt. 

We validate the effectiveness of \neuroprompts by using our framework to produce optimized prompts and images for over 100k baseline prompts. Through automated evaluation, we show that our optimized prompts produce images with significantly higher aesthetics than un-optimized baseline prompts. The optimized prompts produced by our approach even outperform those created by human prompt engineers, demonstrating the ability of our application to unlock the full potential of text-to-image generation models to users without any expertise in prompt engineering.



\section{\neuroprompts Framework}


Given an un-optimized prompt provided by a user, which we denote as $x_{u}$, our \neuroprompts framework generates an optimized prompt $x_{o}$ 
to increase the likelihood that text-to-image diffusion models produce an aesthetically-pleasing image when prompted with $x_{o}$. We specifically consider the case where $x_{u}$ is the prefix of $x_{o}$ and produce the enhanced prompt via a two-stage approach. First, we adapt a language model (LM) to produce a text which is steered towards the style of prompts produced by human prompt engineers. We then generate enhanced prompts via our steered LM using a constrained text decoding algorithm (NeuroLogic), which enables user customizability and improves the coverage of image enhancement keywords.

\subsection{LM Adaptation for Prompt Enhancement}

To adapt LMs for prompt engineering, we use a combination of supervised fine-tuning followed by reinforcement learning via the PPO algorithm.

\subsubsection{Supervised fine-tuning (SFT)}

First, we fine-tune a pre-trained LM to adapt the LM's generated text to the style of language commonly used by human prompt engineers. 
We use a pre-trained GPT-2 LM throughout this work due to its demonstrated exceptional performance in natural language processing tasks. However, our framework is broadly compatible with any autoregressive LM.
To fine-tune the LM, we use a large corpus of human-created prompts for text-to-image models, which we describe subsequently in Section~\ref{sec:dataset}. 

\subsubsection{Reinforcement Learning via PPO}
\label{sec:ppo}

Following SFT, we further train our LM by formulating a reward model based on predicted human preferences of images generated by enhanced prompts. We then use our reward model to further train the LM via the PPO algorithm. 

\paragraph{Extracting prefixes from human prompts}

In order to emulate the type of prompts that a non-expert user might enter into our application for enhancement, we created a dataset of un-optimized prompts which is derived from human-authored prompts.
Human prompt engineers commonly optimize prompts by adding a comma-separated list of keywords describing artists, styles, vibes, and other artistic attributes at the end of the prompt.
Thus, we truncate each of the human-authored prompts in our training dataset to contain only the substring prior to the first occurrence of a comma.
We refer to the resulting prompts as \textit{prefixes}.

\paragraph{Image generation with Stable Diffusion}

Let $x_{u}$ hereafter denote a prompt prefix, which we utilize as a proxy for an un-optimized prompt provided by a user. For each $x_{u}$ derived from our training dataset, we create a corresponding optimized prompt $x_{o}$ using our SFT-trained LM.
Given the prefix, the SFT model generates a continuation of it, leveraging the prompt distribution it has learned from the training dataset (e.g., incorporating modifiers).
We employ beam search with a beam size of 8 and a length penalty of 1.0 for this stage of SFT. We then use Stable Diffusion to generate images $y_{u}$ and $y_{o}$ for prompts $x_{u}$ and $x_{o}$, respectively.

\paragraph{Reward modeling (RM)}
\label{sec:reward}
We evaluate the effectiveness of our SFT LM at optimizing prompts using PickScore \cite{lu-etal-2021-neurologic}, a text-image scoring function for predicting user preferences.
PickScore was trained on the Pick-a-Pic dataset, which contains over 500k text-to-image prompts, generated images, and user-labeled preferences.

PickScore utilizes the architecture of CLIP; given a prompt $x$ and an image $y$, the scoring function $s$ computes a $d$-dimensional vector representation of $x$ and $y$ using a text and image decoder (respectively), returning their inner product:
\begin{equation}
g_{pick}(x, y) = E_{txt}(x) \cdot E_{img}(y)^{T}
\end{equation}
where $g_{pick}(x, y)$ denotes the score of the quality of a generated image $y$ given the prompt $x$.
A higher PickScore indicates a greater likelihood that a user will prefer image $y$ for prompt $x$.

\paragraph{Reinforcement learning (RL)}
We further train our LM using PPO \cite{schulman2017proximal}.
Given the images generated previously for the optimized prompt and prompt prefix, we use PPO to optimize the reward determined by the PickScore:
\begin{equation*} 
\resizebox{1\columnwidth}{!}{
\label{eq1}
$\operatorname{R}\left(x,y \right)=E_{\left(x, y_{u}, y_{o}\right) \sim D}[g_{pick}\left(x, y_{o}\right)-g_{pick}\left(x, y_{u}\right)]$
}
\end{equation*}
where \( g_{pick}(x, y) \) is the scalar output of the PickScore model for prompt \( x \) and image \( y \), \(y_{u}\) is the image generated from the un-optimized prompt, \(y_{o}\) is the image generated from the optimized prompt, and $D$ is the dataset.
This phase of training with PPO further adapts the LM by taking into consideration the predicted human preferences for images generated by the optimized prompts.

\subsection{Constrained Decoding via NeuroLogic}
\label{sec:neurologic}

After training our LM via SFT and PPO, we generate enhanced prompts from it at inference time using NeuroLogic Decoding \cite{lu-etal-2021-neurologic}. NeuroLogic is a constrained text decoding algorithm that enables control over the output of autoregressive LMs via lexical constraints. Specifically, NeuroLogic generates text satisfying a set of clauses $\{C_{i} \mid i \in 1, \cdots m\}$ consisting of one or more predicates specified in conjunctive normal form:
\begin{equation*}
    \underbrace{(D_{1} \lor D_{2} \cdots \lor D_{i})}_\textrm{$C_1$} \land \cdots \land \underbrace{(D_{k} \lor D_{k+1} \cdots \lor D_{n})}_\textrm{$C_m$}
\end{equation*}
where $D_{i}$ is a predicate representing a constraint $D(\mathbf{a}_i, \mathbf{y})$ which evaluates as true if the subsequence $\mathbf{a}_i$ appears in the generated sequence $\mathbf{y}$. NeuroLogic also supports negation of predicates (i.e., $\lnot D_{i}$), specifying the minimum and/or maximum number of predicates within a clause which can be used to satisfy it, and enforcement of clause satisfaction order \cite{howard2023neurocomparatives}. 



We use a curated set of prompt enhancement keywords\footnote{From \href{https://docs.google.com/spreadsheets/d/1-snKDn38-KypoYCk9XLPg799bHcNFSBAVu2HVvFEAkA}{prompt engineering templates}} to formulate clauses which must be satisfied in the optimized prompt. Specifically, we create six clauses consisting of keywords for styles, artists, formats, perspectives, boosters, and vibes (see \autoref{tab:all_clusters} of Appendix~\ref{ap:clusters} for details). Each clause is satisfied when the generated sequence contains one of the keywords from each category. By default, a clause contains five randomly sampled keywords from its corresponding category. However, our application allows users to manually specify which keywords can satisfy each clause to provide more fine-grained control over the optimized prompt. 

\section{Experiments}
\label{sec:experiments}

\begin{table}[t]
 \small 
  \centering
    \begin{tabular}{lr}
    \toprule
    \textbf{Model} & \multicolumn{1}{r}{\textbf{Aesthetics Score}} \\
    \midrule
    Original prefix & 5.64 \\
    Original (human) prompt & 5.92  \\    
    \midrule
    SFT only & 6.02  \\ 
    \neuroprompts w/o PPO & 6.05 \\
    \neuroprompts w/o NeuroLogic  & 6.22 \\
    \neuroprompts & 6.27 \\    

    \bottomrule
    \end{tabular}%
      \caption{Aesthetics scores calculated for images generated by \neuroprompts and baseline methods}
  \label{tab:keystats}%
\end{table}%


\subsection{Dataset}
\label{sec:dataset}
For supervised fine-tuning and reinforcement learning, we utilize the DiffusionDB dataset \cite{wang2022diffusiondb}, a large dataset of human-created prompts.
In the reinforcement learning stage, we truncate the prompt to contain only the substring before the first occurrence of a comma, as previously described in Section~\ref{sec:ppo}. This allows for improved exploration of paraphrasing (see App.~\ref{ap:dataset} for details).

\begin{figure*}[hbt!]
	\centering
	\includegraphics[width=1\textwidth]{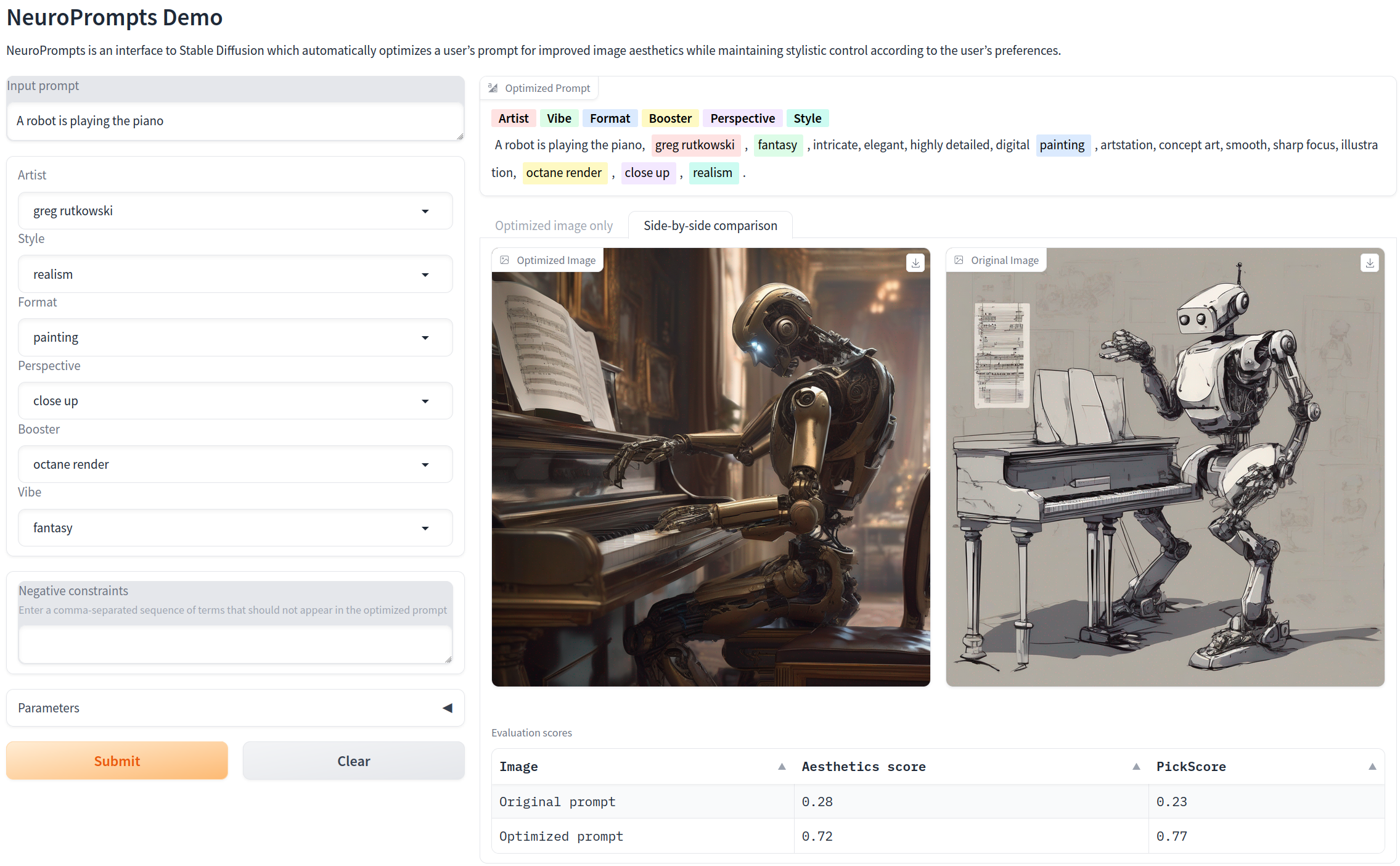}
	\caption{The interface of \neuroprompts in side-by-side comparison mode}
	\label{fig:interface}
\end{figure*}

\subsection{Experimental setting} 

To adapt GPT-2 to the style of prompts created by human prompt engineering, we train it on 600k prompts sampled from DiffusionDB. Specifically, we fine-tune the model for 15,000 steps with a learning rate of 5e-5 and batch size of 256.
We then further train our SFT LM with PPO for 10k episodes using a batch size of 128, a minibatch size of one, four PPO epochs per batch, and a constant learning rate of 5e-5. We used a value loss coefficient of 0.1 and a KL reward coefficient of 0.2. This stage of training was conducted using the PPO implementation from \citep{vonwerra2022trl}.

We use two metrics to evaluate the benefits of our prompt adaptation for text-to-image models: aesthetics score and PickScore.
Aesthetics score is a measure of the overall quality of the generated image and is computed by a model\footnote{We use \href{https://github.com/christophschuhmann/improved-aesthetic-predictor}{Improved Aesthetic Predictor}} trained on LAION \cite{schuhmann2022laion5b} which predicts the likelihood that a human would find the image aesthetically pleasing.
As detailed in Section~\ref{sec:reward}, PickScore measures how likely a human would prefer the generated image using a fine-tuned clip model.
We use a different set of 100k prompts (non-overlapping with our 600k training set) sampled from DiffusionDB for this evaluation and compare the performance of our prompt optimization method to three baselines: (1) the original human-authored prompt from DiffusionDB; (2) the prefix extracted from human-authored prompts, which we consider a proxy for user-provided prompts; and (3) prompts enhanced only using our LM trained with supervised fine-tuning (i.e., without PPO training).

\subsection{Results} 

\paragraph{Optimized prompts produce images with higher aesthetics score}

\autoref{tab:keystats} provides the mean aesthetic scores of images produced by our optimized prompts as well as other baseline methods. \neuroprompts outperforms all other baselines, achieving an average aesthetics score of 6.27, which is an absolute improvement of 0.63 over images produced by un-optimized prompt prefixes. \neuroprompts even outperform human-authored prompts by a margin of 0.35, which could be attributed to how our method learns the relationship between prompt enhancement keywords and image aesthetics across a large dataset of human-authored prompts.
These results demonstrate our framework's effectiveness at generating prompts that produce aesthetically pleasing images.

To analyze the impact of different components of our framework, \autoref{tab:keystats} provides results for variations without PPO training and constrained decoding. PPO training significantly outperforms approaches that only utilize our SFT LM, improving the aesthetics score by approximately 0.2 points. Constrained decoding with NeuroLogic further improves the aesthetics of our PPO-trained model by 0.05, which could be attributed to greater coverage of prompt enhancement keywords. Beyond improvements in aesthetics score, NeuroLogic also enables user control over prompt enhancement.

\paragraph{Optimized prompts achieve higher PickScores}
We further investigated the effect of \neuroprompts on the predicted PickScore of generated images.
Specifically, for each prompt in our DiffusionDB evaluation set, we calculated the PickScore using images generated for the prompt prefix and our optimized prompt. 
Our optimized prompts consistently achieve a higher PickScore than prompt prefixes, with \neuroprompts having an average PickScore of 60\%. This corresponds a 20\% absolute improvement in the predicted likelihood of human preference for our optimized images relative to those produced by prompt prefixes.

\paragraph{Discussion}
Our experiments demonstrate that \neuroprompts consistently produce higher-quality images, indicating that our framework can be used as a practical tool for artists, designers, and other creative professionals to generate high-quality and personalized images without requiring specialized prompt engineering expertise.

\section{\neuroprompts}

The user interface of \neuroprompts is depicted in \autoref{fig:interface}. The application's inputs include the initial prompt as well as selection fields for specifying the clauses used to populate constraints for style, artist, format, booster, perspective, and vibe. Additionally, a negative constraints input allows the user to specify one or more phrases which should be excluded from the optimized prompt. While the initial prompt is required, all other fields are optional; if left unselected, clauses for each constraint set will be automatically populated as described previously in Section~\ref{sec:neurologic}. This functionality allows the user to take control of the constrained generation process if desired or simply rely on our framework to optimize the prompt automatically. 

After clicking the submit button, the optimized prompt is displayed at the top of the screen. If constraints were selected by the user, the optimized prompt will appear with color-coded highlighting to show where each constraint has been satisfied in the generated sequence. The image produced by Stable Diffusion for the optimized prompt is displayed directly below the optimized prompt in the center of the interface. If the user selects the side-by-side comparison tab, an image generated for the original prompt is also displayed to the right of the optimized image. Additionally, the application calculates PickScore and a normalized aesthetics score for the two images, which is displayed in a table below the images. This side-by-side comparison functionality allows the user to directly assess the impact of our prompt optimizations on the quality of images generated by Stable Diffusion.


\paragraph{Examples of images generated from original and optimized prompts}

To further illustrate the impact of \neuroprompts on image quality, \autoref{tbl:text_image_pair} provides examples of images generated from original prompts and our optimized prompts. 
Each row of the table provides an original (un-optimized) prompt along with images generated by Stable Diffusion for the original prompt (center) and an optimized prompt produced by \neuroprompts (right).
These examples illustrate how \neuroprompts consistently produces a more aesthetically-pleasing image than un-optimized prompts.

\begin{table*}
\centering
\resizebox{1\textwidth}{!}{%
\begin{tabular}{p{0.1cm} c c}
\toprule
\textbf{} & \textbf{Original Image} & \textbf{\neuroprompts Optimized Image} \\
\midrule
\rotatebox[origin=c]{90}{A boy on a horse} & \begin{minipage}{0.5\textwidth} \centering \includegraphics[width=1\textwidth]{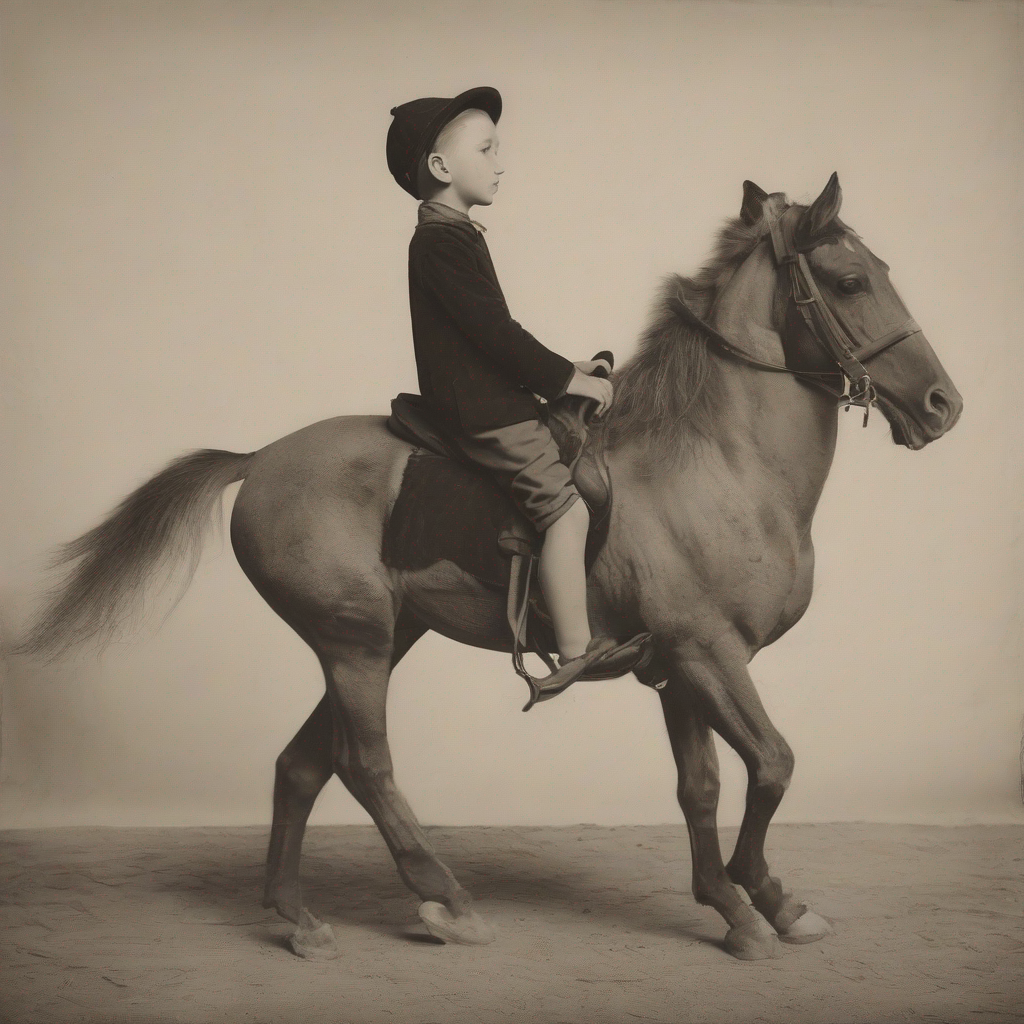} \end{minipage} & \begin{minipage}{0.5\textwidth} \centering\includegraphics[width=1\textwidth]{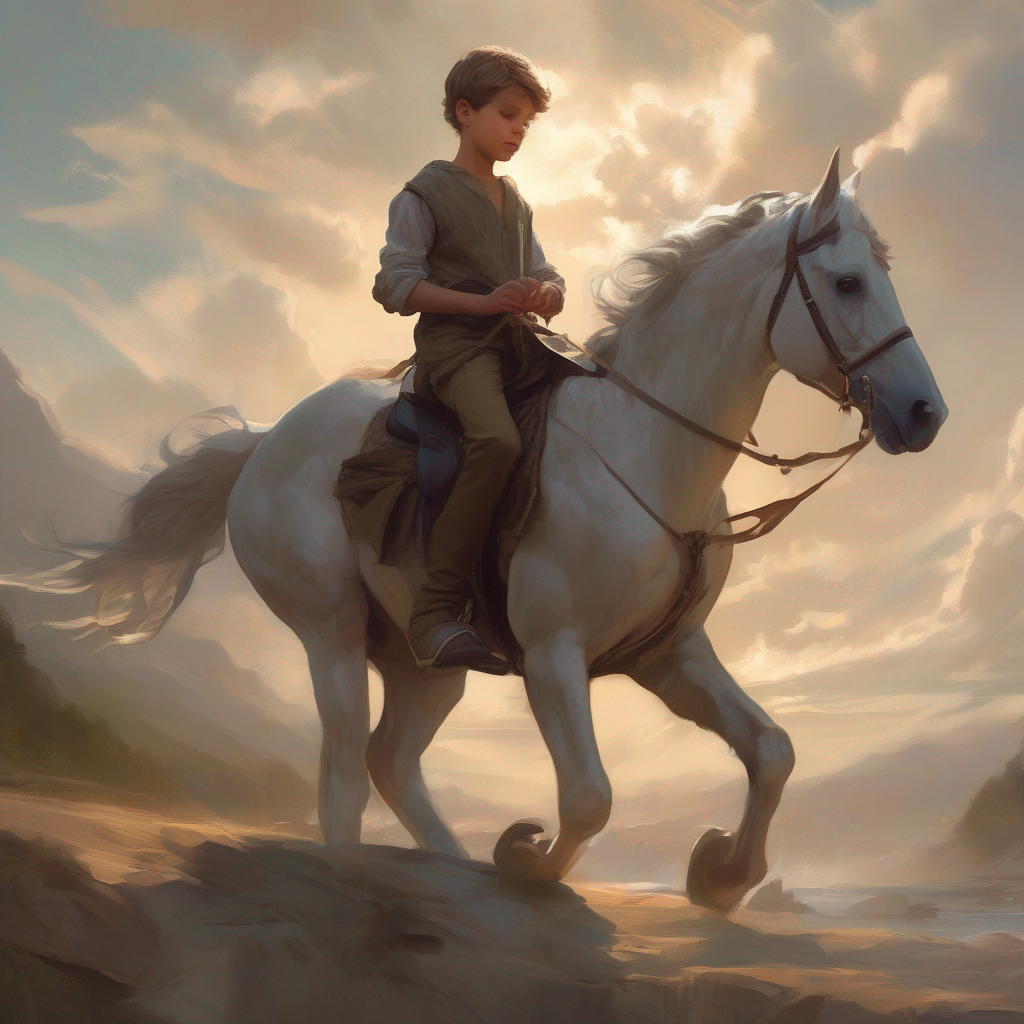} \end{minipage} \\
\midrule
\rotatebox[origin=c]{90}{A tropical beach with palm trees} & \begin{minipage}{0.5\textwidth} \centering \includegraphics[width=1\textwidth]{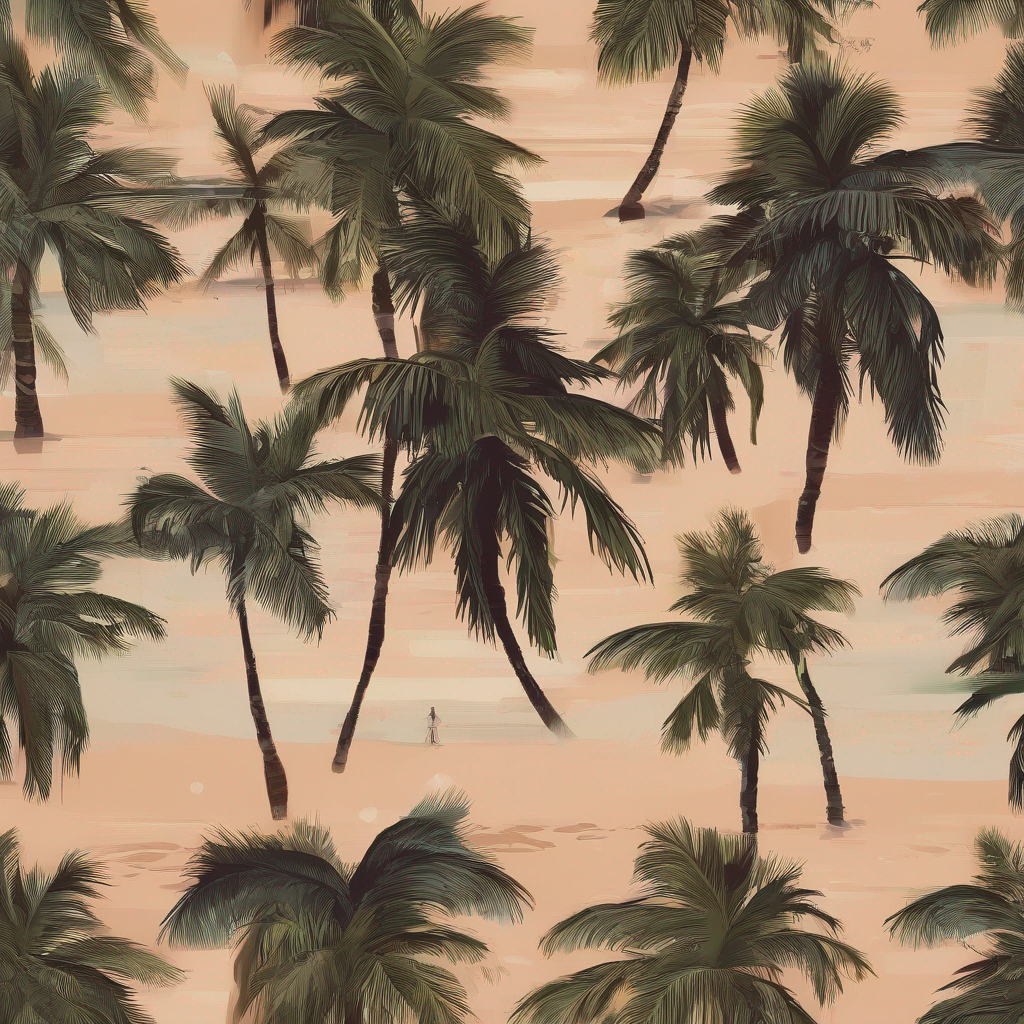} \end{minipage} & \begin{minipage}{0.5\textwidth} \centering\includegraphics[width=1\textwidth]{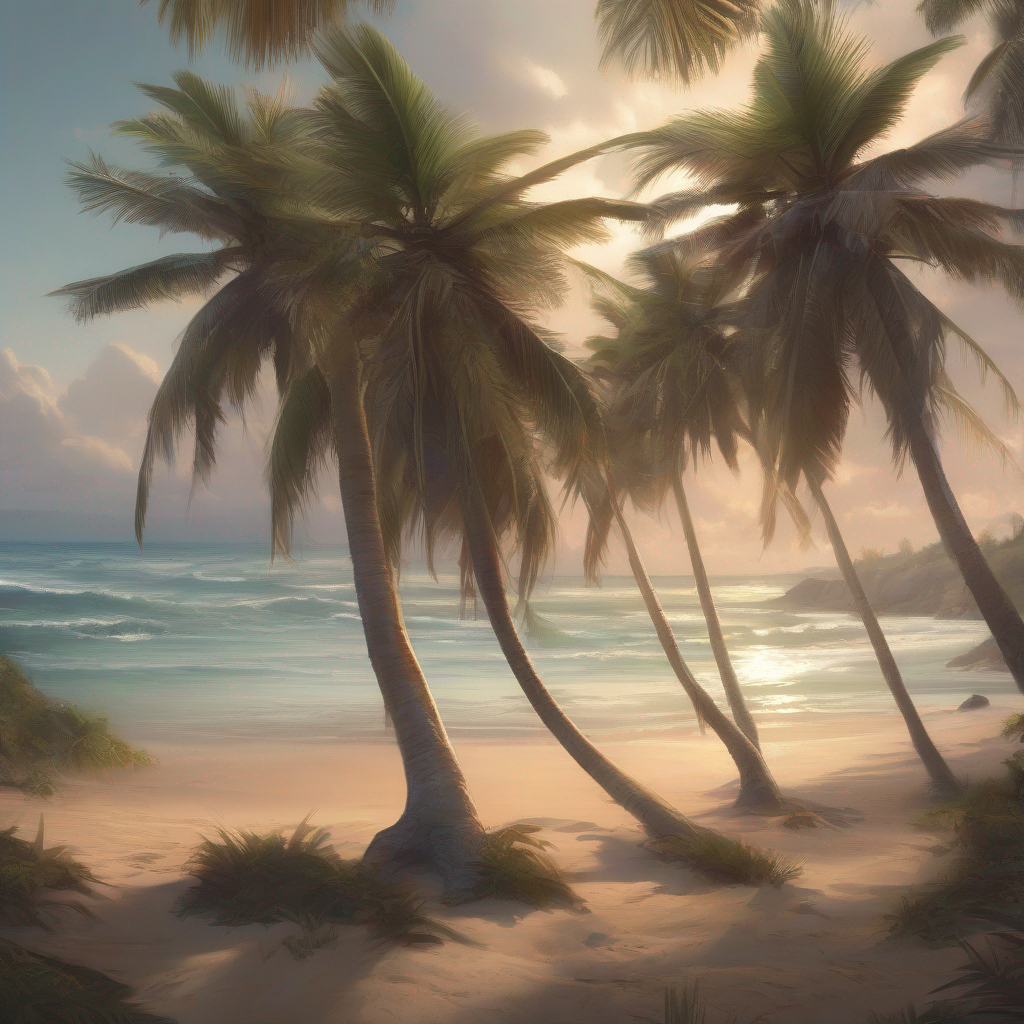} \end{minipage} \\
\midrule
\rotatebox[origin=c]{90}{Two women working in a kitchen} & \begin{minipage}{0.5\textwidth} \centering \includegraphics[width=1\textwidth]{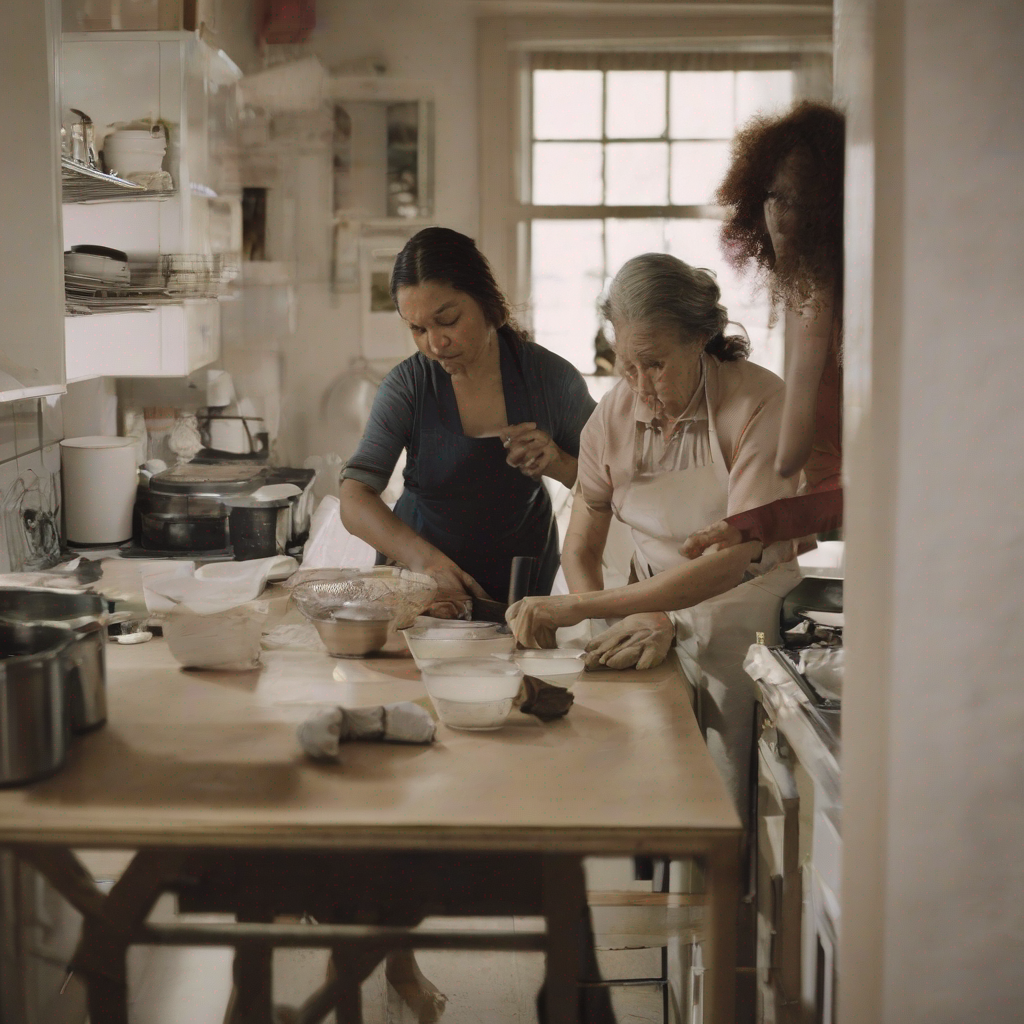} \end{minipage} & \begin{minipage}{0.5\textwidth} \centering\includegraphics[width=1\textwidth]{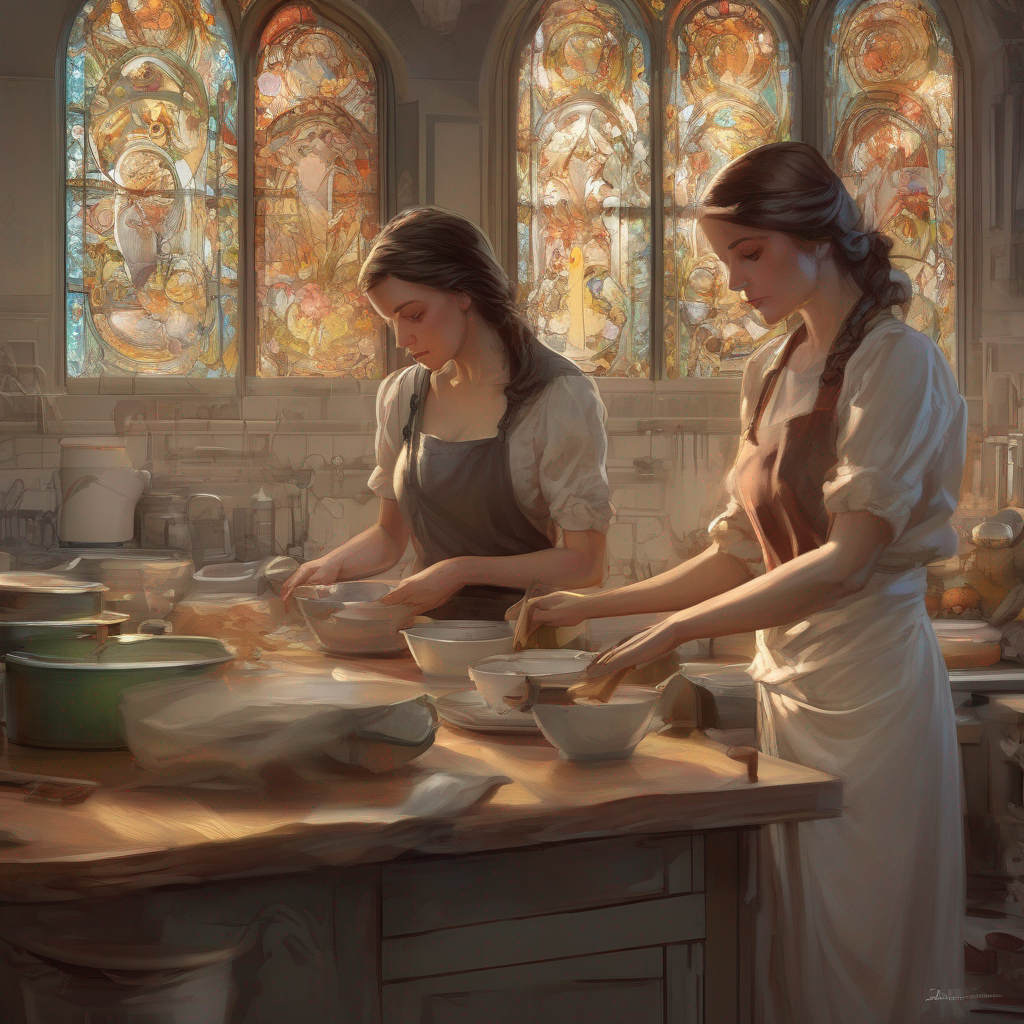} \end{minipage} \\
\bottomrule
\end{tabular}}
\caption{Examples of images generated from original prompts and our optimized prompts. The original (un-optimized) prompt is shown in rotated text to the left of each image pair}
\label{tbl:text_image_pair}
\end{table*}

\section{Related Work}

\paragraph{Text-to-image generation.}
Recent advances in text-to-image generation have led to the release of a variety of models which can translate text prompts into high quality images, including Glide \citep{nichol2021glide}, DALL-E \citep{ramesh2022hierarchical}, ImageGen \citep{saharia2022photorealistic}, and Stable Diffusion \citep{rombach2022high}. Text-to-image diffusion models such as Stable Diffusion encode text prompts using CLIP \citep{radford2021learning}. Images are then generated via a diffusion process by conditioning on the representation of the text encoding in the latent space of an autoencoder.

\paragraph{Prompt engineering.}
Previous studies have demonstrated the superior performance of models trained on manually designed prefix prompts \cite{brown2020language}.
However, these models are heavily dependent on the prompt components \cite{liu2021makes}. 
Research on text-to-image models has focused on proposing keywords \cite{oppenlaender2022taxonomy} and design guidelines \cite{liu2022design}.
Additionally, prior studies have explored the enhancement of LM prompts through differentiable tuning of soft prompts \cite{lester2021power, qin2021learning}.
Similar to our approach, \citet{hao2022optimizing} proposed an automatic prompt engineering scheme via reinforcement learning. In contrast to this prior work, \neuroprompts preserves user interpretabilty and control over the prompt optimization process via the use of symbolic constraints. 

\paragraph{Learning from human preference.}
Human feedback has been used to improve various machine learning systems, and several recent investigations into reinforcement learning from human feedback (RLHF) have shown encouraging outcomes in addressing machine learning challenges. 
These studies include applications to instruction following \cite{ouyang2022training}, summarization \cite{stiennon2020learning} and text-to-image models \cite{lee2023aligning}. While \citet{hao2022optimizing} also leverage RLHF for the purpose of prompt engineering, our approach uses a different reward function based on human preferences for images (PickScore) while providing user control via constrained decoding.

\paragraph{NeuroLogic Decoding}
NeuroLogic Decoding \cite{lu-etal-2021-neurologic} has been extended and applied to various use cases, including A* search \cite{lu2021neurologic} counterfactual generation \cite{howard2022neurocounterfactuals}, inductive knowledge distillation \cite{bhagavatula2022i2d2}, and the acquisition of comparative knowledge \cite{howard2023neurocomparatives}. To the best of our knowledge, our work is the first to explore the applicability of constrained text generation with NeuroLogic to prompt optimization. 

\section{Conclusion}
We presented \neuroprompts, an application which automatically optimizes user prompts for text-to-image generation. 
\neuroprompts unlocks the full potential of text-to-image diffusion models to users without requiring any training in how to construct an optimal prompt for the model. Therefore, we expect it to increase the accessibility of such models while improving their ability to be deployed in a more automated fashion. In future work, we would like to extend \neuroprompts to video generation models and other settings which can benefit from automated prompt engineering. 

\section*{Limitations}

While \neuroprompts is broadly compatible with any text-to-image generation model, we only evaluated its use with Stable Diffusion in this work due to limited computational resources. Images generated from Stable Diffusion have been shown to exhibit societal biases \cite{luccioni2023stable}; therefore, it is expected that images generated using \neuroprompts will also exhibit similar biases. The automated nature of our prompt enhancement and image generation framework introduces the possibility of content being generated which may be considered offensive or inappropriate to certain individuals. Consequently, user discretion is advised when interacting with \neuroprompts.




\bibliography{anthology,custom}

\begin{thebibliography}{25}
\expandafter\ifx\csname natexlab\endcsname\relax\def\natexlab#1{#1}\fi

\bibitem[{Bhagavatula et~al.(2022)Bhagavatula, Hwang, Downey, Bras, Lu, Sakaguchi, Swayamdipta, West, and Choi}]{bhagavatula2022i2d2}
Chandra Bhagavatula, Jena~D Hwang, Doug Downey, Ronan~Le Bras, Ximing Lu, Keisuke Sakaguchi, Swabha Swayamdipta, Peter West, and Yejin Choi. 2022.
\newblock I2d2: Inductive knowledge distillation with neurologic and self-imitation.
\newblock \emph{arXiv preprint arXiv:2212.09246}.

\bibitem[{Brown et~al.(2020)Brown, Mann, Ryder, Subbiah, Kaplan, Dhariwal, Neelakantan, Shyam, Sastry, Askell et~al.}]{brown2020language}
Tom Brown, Benjamin Mann, Nick Ryder, Melanie Subbiah, Jared~D Kaplan, Prafulla Dhariwal, Arvind Neelakantan, Pranav Shyam, Girish Sastry, Amanda Askell, et~al. 2020.
\newblock Language models are few-shot learners.
\newblock \emph{Advances in neural information processing systems}, 33:1877--1901.

\bibitem[{Hao et~al.(2022)Hao, Chi, Dong, and Wei}]{hao2022optimizing}
Yaru Hao, Zewen Chi, Li~Dong, and Furu Wei. 2022.
\newblock Optimizing prompts for text-to-image generation.
\newblock \emph{arXiv preprint arXiv:2212.09611}.

\bibitem[{Howard et~al.(2022)Howard, Singer, Lal, Choi, and Swayamdipta}]{howard2022neurocounterfactuals}
Phillip Howard, Gadi Singer, Vasudev Lal, Yejin Choi, and Swabha Swayamdipta. 2022.
\newblock Neurocounterfactuals: Beyond minimal-edit counterfactuals for richer data augmentation.
\newblock \emph{arXiv preprint arXiv:2210.12365}.

\bibitem[{Howard et~al.(2023)Howard, Wang, Lal, Singer, Choi, and Swayamdipta}]{howard2023neurocomparatives}
Phillip Howard, Junlin Wang, Vasudev Lal, Gadi Singer, Yejin Choi, and Swabha Swayamdipta. 2023.
\newblock Neurocomparatives: Neuro-symbolic distillation of comparative knowledge.
\newblock \emph{arXiv preprint arXiv:2305.04978}.

\bibitem[{Lee et~al.(2023)Lee, Liu, Ryu, Watkins, Du, Boutilier, Abbeel, Ghavamzadeh, and Gu}]{lee2023aligning}
Kimin Lee, Hao Liu, Moonkyung Ryu, Olivia Watkins, Yuqing Du, Craig Boutilier, Pieter Abbeel, Mohammad Ghavamzadeh, and Shixiang~Shane Gu. 2023.
\newblock Aligning text-to-image models using human feedback.
\newblock \emph{arXiv preprint arXiv:2302.12192}.

\bibitem[{Lester et~al.(2021)Lester, Al-Rfou, and Constant}]{lester2021power}
Brian Lester, Rami Al-Rfou, and Noah Constant. 2021.
\newblock The power of scale for parameter-efficient prompt tuning.
\newblock \emph{arXiv preprint arXiv:2104.08691}.

\bibitem[{Liu et~al.(2021)Liu, Shen, Zhang, Dolan, Carin, and Chen}]{liu2021makes}
Jiachang Liu, Dinghan Shen, Yizhe Zhang, Bill Dolan, Lawrence Carin, and Weizhu Chen. 2021.
\newblock What makes good in-context examples for gpt-$3 $?
\newblock \emph{arXiv preprint arXiv:2101.06804}.

\bibitem[{Liu and Chilton(2022)}]{liu2022design}
Vivian Liu and Lydia~B Chilton. 2022.
\newblock Design guidelines for prompt engineering text-to-image generative models.
\newblock In \emph{Proceedings of the 2022 CHI Conference on Human Factors in Computing Systems}, pages 1--23.

\bibitem[{Lu et~al.(2021{\natexlab{a}})Lu, Welleck, West, Jiang, Kasai, Khashabi, Bras, Qin, Yu, Zellers et~al.}]{lu2021neurologic}
Ximing Lu, Sean Welleck, Peter West, Liwei Jiang, Jungo Kasai, Daniel Khashabi, Ronan~Le Bras, Lianhui Qin, Youngjae Yu, Rowan Zellers, et~al. 2021{\natexlab{a}}.
\newblock Neurologic a* esque decoding: Constrained text generation with lookahead heuristics.
\newblock \emph{arXiv preprint arXiv:2112.08726}.

\bibitem[{Lu et~al.(2021{\natexlab{b}})Lu, West, Zellers, Le~Bras, Bhagavatula, and Choi}]{lu-etal-2021-neurologic}
Ximing Lu, Peter West, Rowan Zellers, Ronan Le~Bras, Chandra Bhagavatula, and Yejin Choi. 2021{\natexlab{b}}.
\newblock \href {https://doi.org/10.18653/v1/2021.naacl-main.339} {{N}euro{L}ogic decoding: (un)supervised neural text generation with predicate logic constraints}.
\newblock In \emph{Proceedings of the 2021 Conference of the North American Chapter of the Association for Computational Linguistics: Human Language Technologies}, pages 4288--4299, Online. Association for Computational Linguistics.

\bibitem[{Luccioni et~al.(2023)Luccioni, Akiki, Mitchell, and Jernite}]{luccioni2023stable}
Alexandra~Sasha Luccioni, Christopher Akiki, Margaret Mitchell, and Yacine Jernite. 2023.
\newblock Stable bias: Analyzing societal representations in diffusion models.
\newblock \emph{arXiv preprint arXiv:2303.11408}.

\bibitem[{Nichol et~al.(2021)Nichol, Dhariwal, Ramesh, Shyam, Mishkin, McGrew, Sutskever, and Chen}]{nichol2021glide}
Alex Nichol, Prafulla Dhariwal, Aditya Ramesh, Pranav Shyam, Pamela Mishkin, Bob McGrew, Ilya Sutskever, and Mark Chen. 2021.
\newblock Glide: Towards photorealistic image generation and editing with text-guided diffusion models.
\newblock \emph{arXiv preprint arXiv:2112.10741}.

\bibitem[{Oppenlaender(2022)}]{oppenlaender2022taxonomy}
Jonas Oppenlaender. 2022.
\newblock A taxonomy of prompt modifiers for text-to-image generation.
\newblock \emph{arXiv preprint arXiv:2204.13988}.

\bibitem[{Ouyang et~al.(2022)Ouyang, Wu, Jiang, Almeida, Wainwright, Mishkin, Zhang, Agarwal, Slama, Ray et~al.}]{ouyang2022training}
Long Ouyang, Jeffrey Wu, Xu~Jiang, Diogo Almeida, Carroll Wainwright, Pamela Mishkin, Chong Zhang, Sandhini Agarwal, Katarina Slama, Alex Ray, et~al. 2022.
\newblock Training language models to follow instructions with human feedback.
\newblock \emph{Advances in Neural Information Processing Systems}, 35:27730--27744.

\bibitem[{Qin and Eisner(2021)}]{qin2021learning}
Guanghui Qin and Jason Eisner. 2021.
\newblock Learning how to ask: Querying lms with mixtures of soft prompts.
\newblock \emph{arXiv preprint arXiv:2104.06599}.

\bibitem[{Radford et~al.(2021)Radford, Kim, Hallacy, Ramesh, Goh, Agarwal, Sastry, Askell, Mishkin, Clark et~al.}]{radford2021learning}
Alec Radford, Jong~Wook Kim, Chris Hallacy, Aditya Ramesh, Gabriel Goh, Sandhini Agarwal, Girish Sastry, Amanda Askell, Pamela Mishkin, Jack Clark, et~al. 2021.
\newblock Learning transferable visual models from natural language supervision.
\newblock In \emph{International conference on machine learning}, pages 8748--8763. PMLR.

\bibitem[{Ramesh et~al.(2022)Ramesh, Dhariwal, Nichol, Chu, and Chen}]{ramesh2022hierarchical}
Aditya Ramesh, Prafulla Dhariwal, Alex Nichol, Casey Chu, and Mark Chen. 2022.
\newblock Hierarchical text-conditional image generation with clip latents.
\newblock \emph{arXiv preprint arXiv:2204.06125}, 1(2):3.

\bibitem[{Rombach et~al.(2022)Rombach, Blattmann, Lorenz, Esser, and Ommer}]{rombach2022high}
Robin Rombach, Andreas Blattmann, Dominik Lorenz, Patrick Esser, and Bj{\"o}rn Ommer. 2022.
\newblock High-resolution image synthesis with latent diffusion models.
\newblock In \emph{Proceedings of the IEEE/CVF conference on computer vision and pattern recognition}, pages 10684--10695.

\bibitem[{Saharia et~al.(2022)Saharia, Chan, Saxena, Li, Whang, Denton, Ghasemipour, Gontijo~Lopes, Karagol~Ayan, Salimans et~al.}]{saharia2022photorealistic}
Chitwan Saharia, William Chan, Saurabh Saxena, Lala Li, Jay Whang, Emily~L Denton, Kamyar Ghasemipour, Raphael Gontijo~Lopes, Burcu Karagol~Ayan, Tim Salimans, et~al. 2022.
\newblock Photorealistic text-to-image diffusion models with deep language understanding.
\newblock \emph{Advances in Neural Information Processing Systems}, 35:36479--36494.

\bibitem[{Schuhmann et~al.(2022)Schuhmann, Beaumont, Vencu, Gordon, Wightman, Cherti, Coombes, Katta, Mullis, Wortsman, Schramowski, Kundurthy, Crowson, Schmidt, Kaczmarczyk, and Jitsev}]{schuhmann2022laion5b}
Christoph Schuhmann, Romain Beaumont, Richard Vencu, Cade Gordon, Ross Wightman, Mehdi Cherti, Theo Coombes, Aarush Katta, Clayton Mullis, Mitchell Wortsman, Patrick Schramowski, Srivatsa Kundurthy, Katherine Crowson, Ludwig Schmidt, Robert Kaczmarczyk, and Jenia Jitsev. 2022.
\newblock \href {http://arxiv.org/abs/2210.08402} {Laion-5b: An open large-scale dataset for training next generation image-text models}.

\bibitem[{Schulman et~al.(2017)Schulman, Wolski, Dhariwal, Radford, and Klimov}]{schulman2017proximal}
John Schulman, Filip Wolski, Prafulla Dhariwal, Alec Radford, and Oleg Klimov. 2017.
\newblock Proximal policy optimization algorithms.
\newblock \emph{arXiv preprint arXiv:1707.06347}.

\bibitem[{Stiennon et~al.(2020)Stiennon, Ouyang, Wu, Ziegler, Lowe, Voss, Radford, Amodei, and Christiano}]{stiennon2020learning}
Nisan Stiennon, Long Ouyang, Jeffrey Wu, Daniel Ziegler, Ryan Lowe, Chelsea Voss, Alec Radford, Dario Amodei, and Paul~F Christiano. 2020.
\newblock Learning to summarize with human feedback.
\newblock \emph{Advances in Neural Information Processing Systems}, 33:3008--3021.

\bibitem[{von Werra et~al.(2020)von Werra, Belkada, Tunstall, Beeching, Thrush, and Lambert}]{vonwerra2022trl}
Leandro von Werra, Younes Belkada, Lewis Tunstall, Edward Beeching, Tristan Thrush, and Nathan Lambert. 2020.
\newblock Trl: Transformer reinforcement learning.
\newblock \url{https://github.com/lvwerra/trl}.

\bibitem[{Wang et~al.(2022)Wang, Montoya, Munechika, Yang, Hoover, and Chau}]{wang2022diffusiondb}
Zijie~J Wang, Evan Montoya, David Munechika, Haoyang Yang, Benjamin Hoover, and Duen~Horng Chau. 2022.
\newblock Diffusiondb: A large-scale prompt gallery dataset for text-to-image generative models.
\newblock \emph{arXiv preprint arXiv:2210.14896}.

\end{thebibliography}
\bibliographystyle{acl_natbib}

\appendix

\section{Appendix}


\subsection{Dataset}\label{ap:dataset}

To train and evaluate our adaptive framework for prompt enhancement in text-to-image generation, we utilized the DiffusionDB dataset \cite{wang2022diffusiondb}, a large dataset of human-created prompts.
We use a subset of 600k prompts from this dataset to conduct supervised fine-tuning of our LM. For the reinforcement learning stage of training, we use a different subset of 400k prompts from DiffusionDB.
For each of the 400k prompts, we truncate the prompt to contain only the substring before the first occurrence of a comma, assuming that modifiers generally appear after the first comma. This approach allows for improved exploration of paraphrasing by our policy.
We filtered examples with a significant overlap between the prefix and the entire prompt. To achieve this, we used a sentence similarity threshold of 0.6 overlap and excluded cases which exceeded this threshold.

\subsection{Prompt enhancement keywords}\label{ap:clusters}

Table~\ref{tab:all_clusters} provides the complete set of prompt enhancement keywords utilized in our constraint sets. 

\begin{table*}
\resizebox{\textwidth}{!}{
\begin{tabular}{llllll}
\toprule
\textbf{Style} & \textbf{Artist} & \textbf{Format} & \textbf{Boosters} & \textbf{Vibes} & \textbf{Perspective} \\ \midrule
	expressionism & pablo picasso & watercolor painting & trending on artstation & control the soul & long shot \\
suminagashi & edvard munch & crayon drawing & octane render & futuristic & plain background \\
surrealism & henri matisse & US patent & ultra high poly & utopian & isometric \\
anime & thomas cole & kindergartener drawing & extremely detailed & dystopian & panoramic \\
art deco & mark rothko & cartoon & very beautiful & blade runner & wide angle \\
photorealism & alphonse mucha & in Mario Kart & studio lighting & cinematic & hard lighting \\
cyberpunk & leonardo da vinci & pixel art & fantastic & fantasy & knolling \\
synthwave & claude monet & diagram & postprocessing & elegant & shallow depth of field \\
realism & james gurney & album art cover & well preserved & magnificent & extreme wide shot \\
pop art & toshi yoshida & under an electron microscope & 4k & retrofuturistic & drone \\
pixar movies & zdzislaw beksinski & photograph & arnold render & awesome & from behind \\
abstract organic & gustave doré & pencil sketch & detailed & transhumanist & landscape \\
dadaism & georges braque & stained glass window & hyperrealistic & bright & 1/1000 sec shutter \\
neoclassicism & bill watterson & advertising poster & rendering & wormhole & from below \\
ancient art & michelangelo & mugshot & vfx & eclectic & head-and-shoulders shot \\
baroque & greg rutkowski & cross-stitched sampler & high detail & epic & from above \\
art nouveau & vincent van gogh & illustration & zbrush & tasteful & oversaturated filter \\
impressionist & caravaggio & pencil and watercolor drawing & 70mm & gorgeous & aerial view \\
symbolism & diego rivera & in Fortnite & hyper realistic & opaque & telephoto \\
hudson river school & dean cornwell & line art & 8k & old & motion blur \\
suprematism & ralph mcquarrie & product photography & professional & lsd trip & 85mm \\
rococo & rené magritte & in GTA San Andreas & beautiful & lo-fi & viewed from behind \\
pointillism & john constable & news crew reporting live & trending on artstation & emo & through a porthole \\
vaporwave & gustave dore & line drawing & stunning & lucid & dark background \\
futurism & jackson pollock & courtroom sketch & contest winner & moody & fisheye lens \\
skeumorphism & hayao miyazaki & on Sesame Street & wondrous & crystal & through a periscope \\
ukiyo-e & lucian freud & wikiHow & look at that detail & melancholy & white background \\
medieval art & johannes vermeer & daguerreotype & highly detailed & cosmos & on canvas \\
corporate memphis & hieronymus bosch & 3d render & 4k resolution & faded & tilted frame \\
minimalism & hatsune miku & modeling photoshoot & rendered in unreal engine & uplight & framed \\
fauvism & utagawa kuniyoshi & one-line drawing & photorealistic & concept art & low angle \\
renaissance & roy lichtenstein & charcoal drawing & blender 3d & atmospheric & lens flare \\
constructivism & yoji shinkawa & captured on CCTV & digital art & dust & close face \\
cubism & craig mullins & painting & vivid & particulate & over-the-shoulder shot \\
memphis design & claude lorrain & macro 35mm photograph & wow & cute & close up \\
romanticism & funko pop & on America's Got Talent & high poly & stormy & extreme close-up shot \\
hieroglyphics & katsushika hokusai & pastel drawing & unreal engine & magical & midshot \\
\bottomrule
\end{tabular}%
}
\caption{Prompt enhancement keywords utilized in constraint sets}
\label{tab:all_clusters}
\end{table*}

\end{document}